\documentclass[letterpaper, 10 pt, conference]{ieeeconf}  

\IEEEoverridecommandlockouts                              

\usepackage{times}
\usepackage{multicol}
\usepackage[bookmarks=true]{hyperref}
\usepackage{amsfonts} 
\usepackage{xcolor}
\usepackage{graphicx}
\usepackage{amsmath}
\usepackage{booktabs} 
\usepackage{colortbl}
\usepackage{tabularray}
\usepackage{multirow}
\usepackage{fontawesome5}
\usepackage{cite}
\usepackage{cancel}
\usepackage{algorithm}
\usepackage{algorithmic}
\usepackage{caption}

\DeclareMathOperator*{\argmin}{arg\,min}

\definecolor{green}{RGB}{11,155,13}

\newcommand{\model}{{{\textsc{phli}}}}
\newcommand{\planner}{{{Dom Planner}}}

\title{\LARGE \textit{Dom, cars don't fly!---Or do they?}\\ In-Air Vehicle Maneuver for High-Speed Off-Road Navigation}

\author{Anuj Pokhrel,  Aniket Datar, and Xuesu Xiao\\
\thanks{All authors are with the Department of Computer Science, George Mason University
        {\tt\footnotesize \{apokhre, adatar, xiao\}@gmu.edu}}}

\begin{document}

\maketitle

\begin{abstract}
When pushing the speed limit for aggressive off-road navigation on uneven terrain, it is inevitable that vehicles may become airborne from time to time. During time-sensitive tasks, being able to fly over challenging terrain can also save time, instead of cautiously circumventing or slowly negotiating through. However, most off-road autonomy systems operate under the assumption that the vehicles are always on the ground and therefore limit operational speed. In this paper, we present a novel approach for in-air vehicle maneuver during high-speed off-road navigation. 
Based on a hybrid forward kinodynamic model using both physics principles and machine learning, our fixed-horizon, sampling-based motion planner ensures accurate vehicle landing poses and their derivatives within a short airborne time window using vehicle throttle and steering commands. We test our approach in extensive in-air experiments both indoors and outdoors, compare it against an error-driven control method, and demonstrate that precise and timely in-air vehicle maneuver is possible through existing ground vehicle controls.  
\end{abstract}


\section{Introduction}
\label{sec::intro}

Off-road navigation presents various challenges that sharply contrast those encountered in on-road or indoor scenarios.
In unstructured off-road environments, robots must detect and avoid obstacles, evaluate the traversability of varied terrain, and continuously adapt to complex vehicle-terrain interactions.
Tackling all these challenges is essential to prevent terminal states that can jeopardize the mission and damage the robot, such as vehicle rollover and getting stuck. 

One particular challenge of off-road navigation is addressing terrain unevenness.  Current state-of-the-art approaches typically rely on perception-based~\cite{castro2023does} traversability estimation~\cite{seo2023learning, frey2023fast, han2024model, pan2024traverse} to avoid uneven terrain or enforce slow speed~\cite{datar2023learning, datar2024terrain, datar2024toward, pokhrel2024cahsor} to prevent catastrophic failures. By circumventing or slowing down on uneven terrain, 
these approaches have yet to push the limits of off-road vehicles' capabilities to quickly traverse through challenging off-road environments. 

\begin{figure}
  \centering
  \includegraphics[width=\columnwidth]{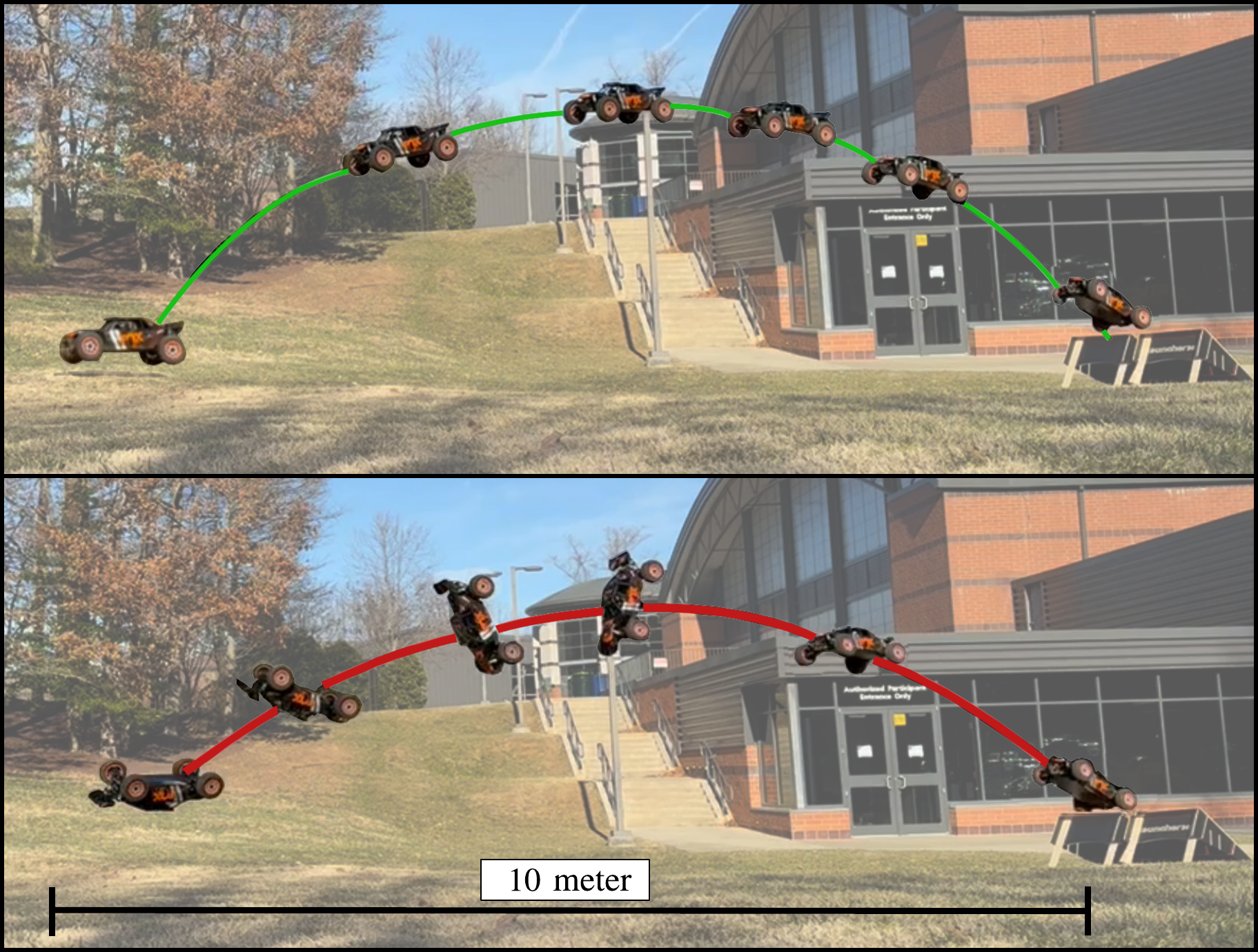}
  \caption{In-air vehicle maneuvers are critical in ensuring safe vehicle landing during high-speed off-road navigation. Top: Precise and timely maneuvers prepare the robot to land with minimal impact. Bottom: Improper maneuvers cause the robot to land on its back, terminating mission execution and risking vehicle damage. }
  \label{fig::flying}
\end{figure}

During time-sensitive off-road missions where achieving the physically feasible maximum speed is necessary, interacting with uneven terrain will cause robots to become airborne from time to time. Additionally, leveraging appropriate terrain structure to take off the vehicle can also efficiently circumvent difficult terrain (in the z direction instead of on the x-y plane) without compromising path length or traversal time. 
However, after air time, deviations from an appropriate landing pose, depending on the receiving terrain geometry (e.g., nose- or tail-down pitch and sideways roll on a flat terrain, or vice versa), can significantly impact landing safety (see examples in Fig.~\ref{fig::flying}).
A critical yet often overlooked aspect of high-speed off-road navigation is the maneuver of a robot during these aerial phases.
Despite limited work on planning before losing ground contact~\cite{lee2023learning}, to the best of our knowledge, no prior work has investigated ground vehicle in-air maneuver to facilitate safe landing.

To this end, we ask the question \textit{how can we control the in-air attitude of a ground robot in a short amount of airborne time only with existing vehicle controls?}
To address this question, we present a novel in-air vehicle attitude planning and control approach by re-purposing existing throttle and steering actions.
Our novel forward kinodynamic model uses the inertial and gyroscopic effects of the spinning wheels to derive the robot's angular accelerations.
These accelerations are used in a Newtonian physics model to calculate the robot's future states. 
Leveraging our model, we also develop a fixed-horizon, sampling-based motion planner specifically designed for quick in-air goal convergence to prepare the vehicle orientation for proper landing. 
Our experiment results on a gimbal platform demonstrate the ability of our method to perform precise and timely in-air maneuvers, showing a significant improvement compared to a traditional error-driven approach.
Furthermore, our outdoor demonstration shows our method's ability to generalize under real-world disturbances like air resistance and uneven weight distribution. 
Our contributions can be summarized as:
~\begin{itemize}
    \item A hybrid dynamics model combining physics principles and data-driven learning for in-air vehicle maneuver driven only by the vehicle throttle and steering actions;
    \item A fixed-horizon, sampling-based motion planner that converges to an appropriate goal state for safe landing;
    \item A set of real-world robot experiments on a gimbal platform and outdoors to demonstrate the effectiveness of our kinodynamic model along with our motion planner. 
\end{itemize}

\section{Related Work}
\label{sec::related}

We review related work in modeling off-road robot dynamics and motion planning for off-road navigation. 

\subsection{Dynamics Modeling for Off-Road Robots}



Accurate dynamics modeling is fundamental to effective robot control, particularly in complex and uncertain off-road environments~\cite{taghavifar2017off}.
Classical wheeled robot dynamics models~\cite{howard2007optimal, hindiyeh2013dynamics}, such as the Ackermann steering and the bicycle model, have demonstrated their utility for conventional ground vehicle navigation through $\mathbb{SE}(2)$ formulations that effectively describe planar motion~\cite{pacejka1992the, rabiee2019friction}.

Off-road navigation demands models to also address the challenges posed by vertically challenging terrain~\cite{datar2023learning, datar2024toward, datar2024terrain,  nazeri2024vertiencoder, nazeri2025vertiformer}, wheel slippage~\cite{yi2009kinematic, ishigami2007path}, and ditch crossing~\cite{han2024model}.
Consequently, data-driven approaches have gained prominence, leveraging sensor data to capture complex dynamics that is difficult to model analytically. 
Yet, purely data-driven methods can be limited by scarce training data and challenges in generalizing across diverse terrain. To address these issues, recent work has focused on physics-informed learning~\cite{kim2022physics, maheshwari2023piaug, cai2025pietra}, which integrates prior physics knowledge into the learning process to regularize model behavior.
Our work also incorporates fundamental physics principles directly into the model structure, thereby combining the benefits of physical consistency with data-driven flexibility. 


When pushing vehicle mobility limits for time-sensitive tasks, high-speed off-road navigation underscores the need for dynamics models operating in $\mathbb{SE}(3)$ as aggressive maneuvers introduce rapid state changes and complex, dynamic vehicle-terrain interactions~\cite{williams2016aggressive, xiao2021learning, karnan2022vi, talia2024demonstrating}. 
Recognizing that safety is essential for high-speed navigation, CAHSOR~\cite{pokhrel2024cahsor} restricts the robot's action to the maximum permissible limit for a given terrain using a $\mathbb{SE}(3)$ model to ensure continuous wheel-terrain contact. Furthermore, other models assume airborne dynamics dictated simply by gravity and therefore pre-plan the aerial trajectory before taking off~\cite{lee2023learning}, but they lack strategies for controlled landing with in-air control. Unlike any prior work, this paper investigates high-speed in-air vehicle maneuver with existing controls in order to safely land with appropriate vehicle configurations.

\subsection{Motion Planning for Off-road Navigation}

Off-road navigation presents unique challenges due to the complex and dynamic nature of the terrain. To address these challenges, a motion planner can take advantage of proprioceptive and exteroceptive sensors to generate traversability estimation~\cite{seo2023scate, frey2023fast, jung2024v, pan2024traverse} and cost maps~\cite{meng2023terrainnet, castro2023does} that can enable online adaptation of control strategies~\cite{siva2019robot, siva2021enhancing} and risk-aware trajectory planning~\cite{cai2022risk, dixit2023step, triest2023learning}. Despite their theoretical optimality, using search-based planners like A* to generate trajectories can be computationally expensive when addressing the non-convex optimization problem in off-road navigation. Error-driven strategies, like PID controllers, are often employed due to their simplicity and low computational overhead. However, when lacking an explicit robot dynamics model, error-driven methods are ill-equipped to tackle the complex and non-linear aggressive maneuvers, especially considering coupled state dimensions and actuation limits.  

In contrast, model-based approaches such as the Model Predictive Path Integral (MPPI) planner~\cite{williams2016aggressive} have been used extensively in off-road navigation~\cite{lee2023learning, datar2024terrain, pokhrel2024cahsor, han2024model}. MPPI leverages sampling to address the non-convex solution space while incorporating robot dynamics and biases its sampling based on the previous best action. With its ability to explore non-convex regions and the use of GPU parallelization~\cite{williams2017model} to accelerate the evaluation of candidate trajectories, it is well-suited for off-road navigation. Nonetheless, those planners do not consider the limited time horizon for in-air maneuvers which is unique and paramount for our problem as an airborne off-road ground robot has a strict and short time window to prepare its configuration for safe landing.

\section{Approach}
\label{sec::approach}

In this section, we first define the in-air vehicle maneuver problem with airborne kinodynamic modeling. Then, we discuss the physics principles of why and how in-air vehicle maneuver is possible through existing vehicle controls, i.e., throttle and steering, based on the inertial and gyroscopic effects of the spinning wheels. 
Motivated by the difficulty in accurately and analytically modeling changing quantities purely based on physics, we present PHysics and Learning based model for In-air vehicle maneuver (\model, pronounced as ``fly''), a precise and efficient hybrid modeling approach. Finally, we introduce our \planner~(\textit{named after Dominic Toretto, who makes cars fly. So does our planner.})  for vehicle trajectory planning to reorient the robot from its current configuration to the goal precisely at when a limited airborne time window expires. 

\subsection{Problem Formulation}
We first formulate a discrete-time forward kinodynamic modeling problem based on a bicycle model, where the subsequent state, $\mathbf{s}_{t+1} \in \mathcal{S}$, is derived from the current state, $\mathbf{s}_t \in \mathcal{S}$, and current action, $\mathbf{a}_t \in \mathcal{A}$, with $\mathcal{S}$ and $\mathcal{A}$ as the state and action spaces respectively. Considering that during aerial phases a conventional ground vehicle does not have control over the three translational components in $\mathbb{SE}(3)$ (which is determined solely by gravity), we include in the vehicle state $\mathbf{s}_t$ a tuple of the three angles in $\mathbb{SO}(3)$, i.e., roll, pitch, and yaw, together with their corresponding angular velocities. We also include rotation per minute of the wheels, $\textrm{rpm}$, and the front steering angle, $\psi$, in the state space, i.e., $\mathcal{S} \subset \mathbb{R}^8$. The state at time $t$ is denoted as 
\begin{equation}
        \mathbf{s}_t = (\textrm{roll}_t, \dot{\textrm{roll}}_t, \textrm{pitch}_t, \dot{\textrm{pitch}}_t, \textrm{yaw}_t, \dot{\textrm{yaw}}_t, \textrm{rpm}_t, \psi_t) \in \mathcal{S}. \nonumber
\end{equation}

The vehicle action, $\mathbf{a}_t$, comprises the rate of change in wheel $\textrm{rpm}$ (throttle) and in steering angle $\psi$, which are common controls available for ground robots. The action in the action space, $\mathcal{A} \subset \mathbb{R}^2$, is thus defined as a two-dimensional tuple:
\begin{equation}
    \begin{gathered}
        \mathbf{a}_t = (\dot{\textrm{rpm}}_t, \dot{\psi}_t) \in \mathcal{A}.
        \nonumber
    \end{gathered}
\end{equation}

A forward kinodynamics model is defined as a function, $f_{\theta}:\mathcal{S} \times \mathcal{A} \rightarrow \mathcal{S}$, parametrized by $\theta$, such that:
\begin{equation}
\mathbf{s}_{t+1} = f_{\theta}(\mathbf{s}_t, \mathbf{a}_t). 
\label{eqn::dynamics}
\end{equation}

After an aerial phase, the vehicle should land on the ground with an appropriate configuration. Unlike traditional navigation planning problems, in which the goal is to achieve a desired state with certain notion of minimal cost (shortest path, lowest energy, etc.), for our in-air maneuver problem the goal is to achieve a goal configuration precisely when a small airborne time window expires, i.e., the landing time, $T$. Therefore, the goal is to achieve: 
\begin{equation}
\mathbf{s}_T^g = (\textrm{roll}_T^g, \dot{\textrm{roll}}_T^g, \textrm{pitch}_T^g, \dot{\textrm{pitch}}_T^g, \textrm{yaw}_T^g, \dot{\textrm{yaw}}_T^g, \textrm{rpm}_T^g, \psi_T^g).  
\end{equation}
Notice that $\mathbf{s}_T^g$ must be achieved precisely at $T$, not later. If it is achieved at $t<T$, the vehicle needs to assure it maintains the same state at the landing time $T$. 
For example, when the landing region is horizontal, $\textrm{roll}_T^g$ and $\textrm{pitch}_T^g$ become zero, while $\dot{\textrm{roll}}_T^g$ and $\dot{\textrm{pitch}}_T^g$ should be as close to zero as possible to minimize impact. $\textrm{yaw}_T^g$ and $\dot{\textrm{yaw}}_T^g$ are often less crucial since they do not cause significant impact during landing. $\textrm{rpm}_T^g$ mostly needs be positive to avoid flipping over the vehicle upon ground contact with forward momentum, and $\psi_T^g$ depends on what the immediate ground maneuver necessary for the vehicle to execute right after landing is, e.g., to quickly swerve to avoid an upcoming obstacle. 

Therefore, the in-air vehicle maneuver problem for safe landing can be formulated as 
\begin{equation}
    \begin{gathered}
    \mathbf{a}_0^*, ..., \mathbf{a}_{T-1}^* = \argmin_{\mathbf{a}_0, ..., \mathbf{a}_{T-1}} \sum_{t=0}^T c(\mathbf{s}_t, \mathbf{s}_T^g),\\
    \textrm{s. t.} \quad \mathbf{s}_{t+1} = f_\theta(\mathbf{s}_t, \mathbf{a}_t), \quad \mathbf{s}_0 \textrm{ is given}, \quad \textrm{and } \mathbf{s}_T = \mathbf{s}_T^g.
    \end{gathered}
\label{eqn::problem}
\end{equation}
$c$ is a state-wise cost function to be minimized, subject to the constraints by the forward kinodynamics, given initial state, and achieving the goal state at the end of the aerial phase $T$. Notice the difference of Problem~\eqref{eqn::problem} to traditional receding-horizon planning problems, where $T$ is a receding horizon which gradually moves towards the goal. Here $T$ is the end of the entire horizon of the problem, i.e., the landing point. 

\begin{figure}
  \centering
  \includegraphics[width=\columnwidth]{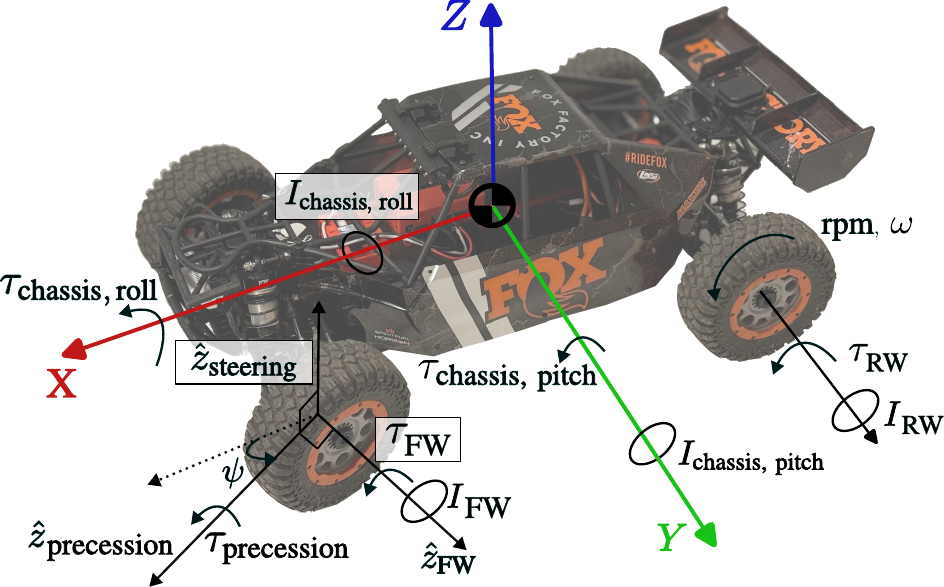}
  \caption{Simplified Bicycle Model with Torques Acting on the Wheels and Chassis due to Wheel Acceleration and Steering.}
  \label{fig::torques}
  \vspace{-9pt}
\end{figure}



\subsection{Physics Principles}
We discuss the physics principles of why and how in-air vehicle maneuver of $\mathbf{s}_{t+1}$ is possible with $\mathbf{a}_t = (\dot{\textrm{rpm}}_t, \dot{\psi}_t)$.

\subsubsection{Inertial Effect}
During aerial phases, the only external force acting on the vehicle is gravity (air resistance is overlooked for simplicity). Therefore, with a constant wheel speed, the vehicle maintains its angular momentum by the principle of conservation of angular momentum. 
All four wheels are rotating at the same speed, because the differentials will not differentiate left and right wheel speed due to a lack of ground contact. Therefore they can be simplified as a front and a rear wheel (FW and RW, bicycle model) with their angular momentum magnitude given by
\begin{equation}
    L_{\textrm{FW/RW}} = I_{\textrm{FW/RW}} \cdot \omega,
    \nonumber
\end{equation}
and direction given by either wheel's spin axis, where $I_{\textrm{FW/RW}}$ is each wheel's moment of inertia along their spin axis and $\omega$ is their angular velocity. 
When the wheels are accelerated by applying a motor torque, equal and opposite reaction torques, $\tau_\textrm{FW} $ and $\tau_\textrm{RW}$, are applied on the chassis by the wheels along their spin axes.
These torques are dependent on the change in angular momentum of the respective wheel $dL_{\textrm{FW/RW}}$:
\begin{equation}
    \tau_{\textrm{FW/RW}} = -\frac{dL_{\textrm{FW/RW}}}{dt}  = -\frac{I_{\textrm{FW/RW}} \cdot d \omega }{dt}= -I_{\textrm{FW/RW}} \cdot \frac{2\pi}{60} \cdot \dot{\textrm{rpm}}. 
    \nonumber
\end{equation}
We can resolve these two torques in individual components, $\tau_\textrm{FW/RW, roll}$ and $\tau_\textrm{FW/RW, pitch}$, aligned with the vehicle's principle axes. The rear wheel's contribution to roll, $\tau_\textrm{RW, roll}$, will always be zero as the rear wheel's spin axis is always perpendicular to the vehicle's roll axis. Its contribution to pitch is $\tau_\textrm{RW, pitch} = \tau_\textrm{RW}$.  For the front wheel turned at a steering angle $\psi$, $\tau_\textrm{FW, roll} = \tau_\textrm{FW} \cdot$ sin($\psi)$ and $\tau_\textrm{FW, pitch} = \tau_\textrm{FW} \cdot $cos($\psi)$. $\tau_\textrm{FW/RW, roll/pitch}$ around the front or real wheel has a resultant torque around the pitch and roll axis passing through the vehicle center of gravity (as the free-flying vehicle body will rotate around the axis with minimum moment of inertia):
\begin{align}
    \tau_{\textrm{chassis, roll}} &= \tau_{\textrm{FW}} \cdot \textrm{sin}(\psi)= -I_{\textrm{FW}}  \cdot \textrm{sin}(\psi) \cdot \frac{2\pi}{60}  \cdot \dot{\textrm{rpm}}, \nonumber\\
     \tau_{\textrm{chassis, pitch}} &= \frac{\tau_{\textrm{FW}} \cdot \textrm{cos}(\psi) + \tau_{\textrm{RW}}}{2} \nonumber \\
     & = -(I_{\textrm{FW}}  \cdot \textrm{cos}(\psi) + I_\textrm{RW}) \cdot \frac{\pi}{60} \cdot \dot{\textrm{rpm}},
     \nonumber
\end{align}
considering the roll axis of the front wheel and of the vehicle chassis align for a bicycle model and assuming the front and rear wheels are equal distance away from the vehicle center of gravity. 
These torque components will cause the vehicle to rotate around the roll and pitch axis based on the chassis moment of inertia around each of these axes, $I_{\textrm{chassis}, \textrm{roll/pitch}}$: 
\begin{equation}
\begin{gathered}
    \ddot{\textrm{roll}} = \frac{\tau_{\textrm{chassis, roll}}}{I_{\textrm{chassis}, \textrm{roll}}} = \frac{- I_\textrm{FW}\cdot  sin(\psi) \cdot 2 \pi }{I_{\textrm{chassis}, \textrm{roll}} \cdot 60 } \cdot \dot{\textrm{rpm}}, \\
    \ddot{\textrm{pitch}} = \frac{\tau_{\textrm{chassis, pitch}}}{I_{\textrm{chassis}, \textrm{pitch}}} = \frac{- (I_\textrm{FW} \cdot cos(\psi) + I_\textrm{RW}) \cdot \pi }{I_{\textrm{chassis}, \textrm{pitch}} \cdot 60 } \cdot \dot{\textrm{rpm}}.
    \label{eqn::roll_dot_dot_1}
\end{gathered}
\end{equation}
We deduce that $\ddot{\textrm{roll}}$ and $\ddot{\textrm{pitch}}$ are proportional to $\dot{\textrm{rpm}}$ and dependent on the $sin$ or $cos$ of the steering angle $\psi$ due to the inertial effect. See Fig.~\ref{fig::torques} for a graphical illustration. 

\subsubsection{Gyroscopic Effect}
Precession describes a phenomenon observed in spinning objects when an external torque is applied perpendicular to the spin axis.
This perpendicular torque causes a change in the axis of rotation and therefore angular momentum to rotate towards the axis of the applied torque while keeping the magnitude of the angular momentum constant.
This rotation of the spin axis is called precession.
For our case of in-air vehicle maneuver, the spinning object is the front wheel, and the external torque perpendicular to the wheel spin axis is from the steering servo to steer the front wheel. 

During in-air maneuvers, the front wheel spinning with angular velocity $\omega$ has angular momentum
\begin{equation}
    L = I_{\textrm{FW}} \cdot \omega \cdot \hat{z}_\textrm{FW}, 
    \nonumber
\end{equation}
where $\hat{z}_\textrm{FW}$ is the unit vector of the front wheel's spin axis. 
When steering the front wheel at a rate $\dot{\psi}$, the spin axis rotates towards the steering axis. This steering motion introduces a precession angular velocity
\begin{equation}
    \Omega = \dot{\psi} \cdot \hat{z}_{\textrm{steering}}, 
    \nonumber
\end{equation}
where $\hat{z}_{\textrm{steering}}$ is the unit vector along the steering axis. 
The reaction torque on the vehicle chassis because of precession is determined by
\begin{equation}
    \tau_{\textrm{precession}} =  L \times \Omega \nonumber = I_{\textrm{FW}} \cdot \omega \cdot \hat{z}_\textrm{FW} \times \dot{\psi} \cdot \hat{z}_{\textrm{steering}}.
    \nonumber
\end{equation}
Since $\hat{z}_\textrm{FW}$ and $\hat{z}_{\textrm{steering}}$ are orthogonal, the magnitude of their cross product is 1 and the resultant direction is $\hat{z}_{\textrm{precession}}$ (see Fig.~\ref{fig::torques}). 
$\tau_{\textrm{precession}}$ has a component on both roll and pitch axes:
\begin{align}
    \tau_{\textrm{precession}, \textrm{roll}} & = cos(\psi) \cdot \tau_{\textrm{precession}} = cos(\psi) \cdot I_{\textrm{FW}}  \cdot \omega \cdot \dot{\psi},\nonumber\\
    \tau_{\textrm{precession}, \textrm{pitch}} & = sin(\psi) \cdot \tau_{\textrm{precession}} = sin(\psi) \cdot I_{\textrm{FW}} \cdot \omega \cdot \dot{\psi}. \nonumber
\end{align}
The $\tau_{\textrm{precession}, \textrm{pitch}}$ component is around the front wheel (away from the pitch axis at vehicle center of gravity), hence the resultant torque $\tau_{\textrm{chassis, pitch}}$ is given by:
\begin{align}
    \tau_{\textrm{chassis, pitch}} &= \frac{\tau_{\textrm{precession,pitch}}}{2},
    \nonumber
\end{align}
as the rear wheel does not contribute to precession.
Similar to the inertial effect, the direction of $\tau_{\textrm{precession}, \textrm{roll}}$ aligns with $\tau_{\textrm{chassis}, \textrm{roll}}$, thus $\tau_{\textrm{chassis}, \textrm{roll}} = \tau_{\textrm{precession}, \textrm{roll}}$. Therefore,
\begin{align}
    \ddot{\textrm{roll}} & = \frac{\tau_{\textrm{chassis}, \textrm{roll}}}{I_{\textrm{chassis, roll}}} = \frac{\tau_{\textrm{precession}, \textrm{roll}}}{I_{\textrm{chassis, roll}}} \nonumber\\
    &= \frac{cos(\psi) \cdot I_{\textrm{FW}} \cdot \omega \cdot \dot{\psi}}{I_{\textrm{chassis, roll}}} 
    = \frac{cos(\psi) \cdot I_{\textrm{FW}} \cdot 2\pi}{I_{\textrm{chassis, roll}}\cdot 60} \cdot \textrm{rpm}\cdot \dot{\psi},\nonumber\\
    \ddot{\textrm{pitch}} &= \frac{\tau_{\textrm{chassis}, \textrm{pitch}}}{I_{\textrm{chassis, pitch}}} = \frac{\tau_{\textrm{precession}, \textrm{pitch}}}{2 \cdot I_{\textrm{chassis, pitch}}} \nonumber\\
    &= \frac{sin(\psi) \cdot I_{\textrm{FW}} \cdot \omega \cdot \dot{\psi}}{2 \cdot I_{\textrm{chassis, pitch}}}
    = \frac{sin(\psi) \cdot I_{\textrm{FW}} \cdot \pi}{I_{\textrm{chassis, pitch}}\cdot 60} \cdot \textrm{rpm}\cdot \dot{\psi}.
    \label{eqn::precession_pitch_roll_dot_dot}
\end{align}
We deduce that $\ddot{\textrm{roll}}$ and $\ddot{\textrm{pitch}}$ are proportional to $\textrm{rpm}$ and $\dot{\psi}$ and dependent on $\psi$ due to the gyroscopic effect. The final $\ddot{\textrm{roll}}$ and $\ddot{\textrm{pitch}}$ are a combined effect of Eqns.~\eqref{eqn::roll_dot_dot_1} and \eqref{eqn::precession_pitch_roll_dot_dot}. 

\subsubsection{Lack of Yaw Control}
In theory, the vehicle action $\mathbf{a}_t = (\dot{\textrm{rpm}}_t, \dot{\psi}_t)$ does not have direct control over the yaw acceleration, $\ddot{\textrm{yaw}}$. But in practice, $\ddot{\textrm{yaw}}$ can be affected by system noises such as air resistance and weight distribution. Furthermore, initial yaw position, $\textrm{yaw}_0$, and yaw rate, $\dot{\textrm{yaw}}_0$, will still affect $\textrm{yaw}_t$ based on its inertia.  

\subsection{\model}
Calculating $\ddot{\textrm{roll}}$ and $\ddot{\textrm{pitch}}$ requires accurate moment of inertia measurements for the vehicle chassis and for the wheels.
These measurements are difficult to analytically derive or precisely measure due to changes in the center of gravity and weight distribution by steering and as a result of suspension movement.
Furthermore, expansion of the tires at high speeds due to centrifugal forces changes the moment of inertia of the wheels as well. 
Hence, we decompose the forward kinodynamics model $\mathbf{s}_{t+1} = f_{\theta}(\mathbf{s}_t, \mathbf{a}_t)$ into two parts, $g_\phi$ and $h_\xi$, i.e., $\mathbf{s}_{t+1} = h_\xi(g_{\phi}(\mathbf{s}_t, \mathbf{a}_t), \mathbf{s}_t, \mathbf{a}_t )$, and then employ a data-driven and a physics-based approach to derive $g_\phi$ and $h_\xi$ respectively. 

$g_\phi$ takes the state and action as input and derives the resultant acceleration of roll, pitch, and yaw
\begin{equation}
    \ddot{\textrm{roll}}_t, \ddot{\textrm{pitch}}_t, \ddot{\textrm{yaw}}_t = g_\phi(\mathbf{s}_t, \mathbf{a}_t),
\end{equation}
where model parameters $\phi$ are learned in a data-driven manner by minimizing a supervised loss. 
The ground truth acceleration values are derived from the recorded Inertial Measurement Unit (IMU) data via differentiation over time.

Once $g_\phi$ is trained, its output angular accelerations are 
fed into $h_\xi$, along with the state $\mathbf{s}_t$ and action $\mathbf{a}_t$. For $\mathbf{s}_{t+1} = h_\xi((\ddot{\textrm{roll}}_t, \ddot{\textrm{pitch}}_t, \ddot{\textrm{yaw}}_t), \mathbf{s}_t, \mathbf{a}_t)$, we have
\begin{equation}
\begin{gathered}
    \dot{\textrm{roll}}_{t+1} = \dot{\textrm{roll}}_{t} + \ddot{\textrm{roll}}_t \cdot dt,\\
    \textrm{roll}_{t+1} = \textrm{roll}_{t} + \dot{\textrm{roll}}_t \cdot dt + \frac{1}{2} \ddot{\textrm{roll}}_t \cdot dt^2,\\
    \textrm{rpm}_{t+1} = \textrm{rpm}_{t} + \dot{\textrm{rpm}}_t \cdot dt,\\
    \psi_{t+1} = \psi_t + \dot{\psi}_t \cdot dt,
    \nonumber
\end{gathered}
\end{equation}
with analogous update equations for pitch and yaw. The only parameter for $h_\xi$ is the integration interval $dt$.  
This completes \model's transition to next state $\mathbf{s}_{t+1}$ given current state $\mathbf{s}_t$ and action $\mathbf{a}_t$ through $g_\phi$ and $h_\xi$. 


\subsection{\planner}
Given the short airtime, we adopt a fixed-horizon, instead of receding-horizon, planning approach, where the horizon is determined by the discretized time remaining until landing.
The planner samples action sequences over this fixed horizon and uses the learned \model~to rollout the corresponding state trajectories.
Each trajectory is then evaluated using a cost function. The planner selects the trajectory with the minimal cost, executes its first action, and replans with the updated (and reduced) remaining time. 
The next planning cycle samples around the best action from the last cycle. 

One critical consideration of \planner~is that, in addition to general physical actuation limits (e.g., limited motor current and torque causing that the actions can never exceed certain thresholds), the under-actuated vehicle controls of $\dot{\textrm{rpm}}$ and $\dot{\psi}$ are further constrained due to state-dependent actuation limits:  
\begin{gather}
 \textrm{rpm}_\textrm{min} - \textrm{rpm}_t \leq \dot{\textrm{rpm}}_t^\textrm{feasible} \leq \textrm{rpm}_\textrm{max} - \textrm{rpm}_t, \nonumber \\
 \psi_\textrm{min} - \psi_t \leq \dot{\psi}_t^\textrm{feasible} \leq \psi_\textrm{max} - \psi_t,
 \nonumber
\end{gather}
which means if $\textrm{rpm}_t$ ($\psi_t$) is at the minimal value, $\dot{\textrm{rpm}}_t^\textrm{feasible}$ ($\dot{\psi}_t^\textrm{feasible}$) cannot be negative, and if $\dot{\textrm{rpm}}_t$ ($\psi_t$) is at the maximal value, $\dot{\textrm{rpm}}_t^\textrm{feasible}$ ($\dot{\psi}_t^\textrm{feasible}$) cannot be positive. 
This limited range of feasible $\dot{\textrm{rpm}}$ and $\dot{\psi}$ reduces possible in-air maneuver options and makes simple error-driven controllers, like PID controllers, inappropriate to enable complex in-air vehicle maneuvers. 

To enable goal convergence within a short aerial phase, the state-wise cost function defined in the fixed-horizon, instead of receding-horizon, problem in Eqn.~\eqref{eqn::problem} is formulated as:
\begin{align}
c(\mathbf{s}_t, \mathbf{s}_T^g) = \sum_{i=1}^K w_i(t) \cdot c_i(\mathbf{s}_t, \mathbf{s}_T^g),
\label{eq::cost_function}
\end{align}
which is a combination of the costs of $K$ state dimensions (in our case, $K=8$). A key difference from conventional state-wise cost function is that each state dimension is dynamically weighed by a weight term $w_i(t)$ as a function of time $t$. For example, for initial time steps, it is more important to quickly align the vehicle angles to prepare for a safe landing pose, while the focus will gradually shift to the angular velocities, wheel rpm, and steering components later on to ensure landing with minimal impact on the vehicle.   


Additionally, based on the sampled trajectory with the lowest cost, \planner~checks its feasibility of reaching the goal state within the time remaining till landing by computing the difference between the final state $\mathbf{s}_T$ and goal state $\mathbf{s}_T^g$. This feasibility check allows the system to either choose a viable trajectory or, if necessary, select an alternate goal to mitigate landing impact.

\section{Implementations}
\label{sec::implementations}

We present the implementation details of our \model~and \planner~onboard a 1/5th-scale vehicle.

\subsection{\model~Implementations}
The learnable function of \model, $g_{\phi}$, is a 4-layer multi-layer perceptron that produces a three-dimensional output of predicted angular accelerations, $(\ddot{\textrm{roll}}_{t+1}, \ddot{\textrm{pitch}}_{t+1}, \ddot{\textrm{yaw}}_{t+1})$.
$g_{\phi}$ is trained to minimize the difference between the predicted angular accelerations and the derived ground truth from IMU.
The outputs of $g_{\phi}$ is fed into $h_\xi$ to calculate the next state of the vehicle $\mathbf{s}_{t+1}$.

\begin{algorithm}
\caption{\planner} \label{alg::planner}
\begin{algorithmic}[1]
   \STATE \textbf{Parameters:} integration interval $dt$ (0.2), sample count $N$ (4000), actuation limits $\lambda$, sampling range $\sigma_{\dot{\textrm{rpm}}}$ (2000) and $\sigma_{\dot{\psi}} (0.2)$, and \model~$f_\theta$
   \STATE \textbf{Input:} Time till landing $T$, initial state $\mathbf{s}_0$, goal state $\mathbf{s}_T^g$, and last best action $(\dot{\textrm{rpm}}_\textrm{best}, \dot{\psi}_\textrm{best})$
    \STATE Compute fixed landing horizon: $H$ = $\frac{T}{dt}$  
    \STATE$\dot{\textrm{rpm}}_{l},\dot{\textrm{rpm}}_{h} = \dot{\textrm{rpm}}_\textrm{best} \pm \sigma_{\dot{\textrm{rpm}}}$ \hfill $\triangleright$ \textit{sampling range for} $\dot{\textrm{rpm}}$
    \STATE$\dot{\psi_{l}}, \dot{\psi_{h}} = \dot{\psi_\textrm{best}} \pm \sigma_{\dot{\psi}}$ \hfill $\triangleright$ \textit{sampling range for} $\dot{\psi}$
     \FOR{($\dot{\textrm{rpm}}_{i}$, $\dot{\psi}_{i}$), $i \in [1, N]$, sampled from range $[\dot{\textrm{rpm}}_{l}$, $\dot{\textrm{rpm}}_{h}]$ and $[\dot{\psi_{l}}, \dot{\psi_{h}}]$}
        \STATE $T_i = \{\mathbf{s}_0\}$
        \STATE $U_i = (\dot{\textrm{rpm}}_{i}, \dot{\psi}_{i})$
        \FOR{$t \in [0, H-1]$}
            \STATE$\mathbf{s}_t$ = \texttt{CLAMP\_STATE}($\mathbf{s}_t$, $\lambda$)
            \STATE$\mathbf{a}_t$ = \texttt{CLAMP\_ACTION}($\dot{\textrm{rpm}_i}, \dot{\psi_i}, \mathbf{s}_t$, $\lambda$) 
            \STATE $\mathbf{s}_{t+1} = f_{\theta}(\mathbf{s}_t, \mathbf{a}_t)$ 
            \STATE $T_i.\textrm{add}(\mathbf{s}_{t+1})$
        \ENDFOR
        \STATE $C_{i}$ = \texttt{CALCULATE\_COST}($T_{i}, \mathbf{s}_T^g$)
    \ENDFOR
    \STATE $T_{\textrm{best}} = T_{\argmin_i(C_i)}$ \hfill $\triangleright$ \textit{minimal-cost trajectory}
    \STATE $U_{\textrm{best}} = U_{\argmin_i(C_i)}$ \hfill $\triangleright$ \textit{best action of this cycle}
    \STATE Return $T_{\textrm{best}}$, $U_{\textrm{best}}$
\end{algorithmic}
\end{algorithm}

\subsection{\planner~Implementation}
\planner's implementation is shown in Algorithm \ref{alg::planner}.
Line 1 first defines \planner's parameters, including integration interval, sample count, actuation limits, sampling range, and \model. Actuation limits include 
$\textrm{rpm}$ rate limit $\dot{\textrm{rpm}}_\textrm{min}$ and $\dot{\textrm{rpm}}_\textrm{max}$ ($\pm5000$), $\textrm{rpm}$ limit $\textrm{rpm}_\textrm{min}$ ($0$) and $\textrm{rpm}_\textrm{max}$ ($1980$), steering rate limit $\dot{\psi}_{\textrm{min}}$ and $\dot{\psi}_{\textrm{max}}$ ($\pm6.5$ rad/s), and steering limit $\psi_\textrm{min}$ and $\psi_\textrm{max}$ ($\pm0.65$ rad).  
Line 2 specifies the algorithm input. Given gravity, time till landing is determined by the initial take-off velocity and terrain geometry. Initial state is the robot state at take-off. Goal state can be defined by the geometry of the receiving terrain. Last best action is the action executed in the last planning cycle. 
The planning cycle is initiated as soon as the vehicle becomes airborne.
The planner calculates the fixed landing horizon $H = \frac{T}{dt}$ by discretizing the time till landing with the interval $dt$ in line 3. 
For each planning cycle, the algorithm uniformly samples 4000 candidate input pairs ($\dot{\textrm{rpm}}_i$, $\dot{\psi}_i$) within the ranges determined by $\sigma_{\dot{\textrm{rpm}}}$ and $\sigma_{\dot{\psi}}$ centered around the last best action $(\dot{\textrm{rpm}}_\textrm{best}, \dot{\psi}_\textrm{best})$ in lines 4 and 5. 
For each sample (line 6), the planner rolls out a trajectory over the horizon $H$ (line 9) using our \model~$f_\theta$ to compute the state transition dynamics (line 12). At every time step, the functions \texttt{CLAMP\_STATE} and \texttt{CLAMP\_ACTION} enforce the state and action limits (lines 10-11). 
After rolling out the full trajectory, \texttt{CALCULATE\_COST} evaluates each trajectory’s cost according to Eqn.~\eqref{eq::cost_function} (line 15). For $w_i(t)$ in Eqn.~\eqref{eq::cost_function}, we simply use higher and lower weights for the angular position and velocity components respectively in the first half of the rollouts, and vice versa in the second half. 
We select the best trajectory corresponding to the minimum cost (line 17) and execute the corresponding action (line 18), The best trajectory and action is returned to initialize the last best action $(\dot{\textrm{rpm}}_\textrm{best}, \dot{\psi}_\textrm{best})$ for the next planning cycle. 
We re-plan at 50 Hz and update the time horizon at every cycle based on the remaining time.

\subsection{Robot, Gimbal, and Dataset}
We implement \model~and \planner~on a 1/5-scale Losi DBXL E2, 4WD Desert Buggy platform, with a top speed of 80+ km/h. The vehicle is equipped with a 9-DoF IMU, Intel RealSense D435i, NVIDIA Jetson Orin NX for perception and planning, two wheel encoders for front and rear wheels respectively, and an Arduino Mega micro-controller for all low-level actuators and the wheel encoders. For simplicity, we only allow the wheels to rotate forward, i.e., $\textrm{rpm} \in [0, \textrm{rpm}_\textrm{max}]$.

\begin{figure}
  \centering
  \includegraphics[width=\columnwidth]{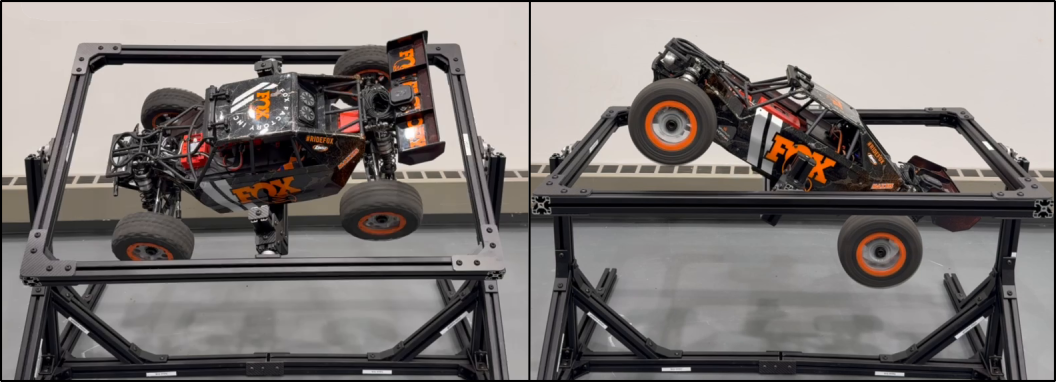}
  \caption{Two-Axis Gimbal for In-Air Dynamics.}
  \label{fig::gimbal}
\end{figure}

To collect a dataset while ensuring the safety of the robot, we construct a 2-axis 1.3 m$\times$1 m$\times$0.65 m aluminum gimbal platform capable of rotating in the roll and pitch axis (Fig.~\ref{fig::gimbal}). 
The purpose of the gimbal is to simulate weightlessness and in-air dynamics.
When mounting the robot on the gimbal with robot's center of gravity aligned to the roll and pitch axis of the gimbal, the robot can freely rotate around the roll and pitch axis. Because existing vehicle controls do not have a direct effect on vehicle yaw acceleration and the gimbal introduces significant extra moment of inertial along the yaw axis, we do not include the yaw angle on the gimbal. 

While the robot is mounted on the gimbal, the robot is commanded with a diverse range of $\dot{\textrm{rpm}}$ and $\dot{\psi}$ to capture all possible ways to change the roll and pitch of the vehicle, including behaviors at the extremes of $\textrm{rpm}$ and steering $\psi$.
The robot configurations during the execution of these commands are recorded and processed to create our tuple of current state, current action, and next state ground truth of the angular accelerations to train $g_\phi$. 
Our dataset includes one hour of gimbal data, split 85/15 for training and validation.

\section{Experiments}
\label{sec::experiments}
Our proposed method is validated through extensive experiments conducted in both indoor and outdoor environments. For practical reasons, the majority of our experiments are performed indoors on the custom-built gimbal, with additional real-world validation provided by outdoor experiments.

\begin{table}
\caption{Quantitative Comparison of \planner~and PID.}
\centering
\setlength{\tabcolsep}{4pt}
\begin{tabular}{ccc}
\toprule
Metric & Dom & PID  \\
\midrule
TT Error (rad) $\downarrow$ &  \textbf{0.23$\pm$0.13} &  0.69$\pm$0.82 \\
TT Completion Time (sec) $\downarrow$  &  \textbf{12.4$\pm$5.60} &  23.7$\pm$14.1\\
TT Success Rate $\uparrow$   &  \textbf{5/5} &  2/5 \\
\midrule
RSC Time (sec) $\downarrow$  &  \textbf{23.3 $\pm$11.2} &  51.2 $\pm$ 16.6\\
RSC Difference (rad) $\downarrow$   & \textbf{0.18 $\pm$0.41} &  0.79 $\pm$ 0.81\\
RSC Success Rate $\uparrow$    &  \textbf{5/5} &  4/5 \\
\midrule
TGR Time Difference (sec) $\downarrow$   &  0.36 $\pm$ 1.80 &  - \\
TGR State Difference (rad) $\downarrow$      &  0.13 $\pm$ 0.29&  - \\
\midrule
SS Correction Time (sec) $\downarrow$   &  \textbf{0.74 $\pm$0.74} &  1.20$\pm$0.98\\
SS Reaction Latency (sec) $\downarrow$   &  \textbf{0.18$\pm$0.13} &  0.65$\pm$0.22\\
\midrule
Outdoor Landing Roll (rad) $\downarrow$   &  0.08 $\pm$ 0.13 &  - \\
Outdoor Landing Pitch (rad) $\downarrow$   &  0.13 $\pm$ 0.21&  -\\
\bottomrule
\end{tabular}%
\label{tab::results}
\end{table}

\begin{figure*}
  \centering
  \includegraphics[width=0.32\textwidth]{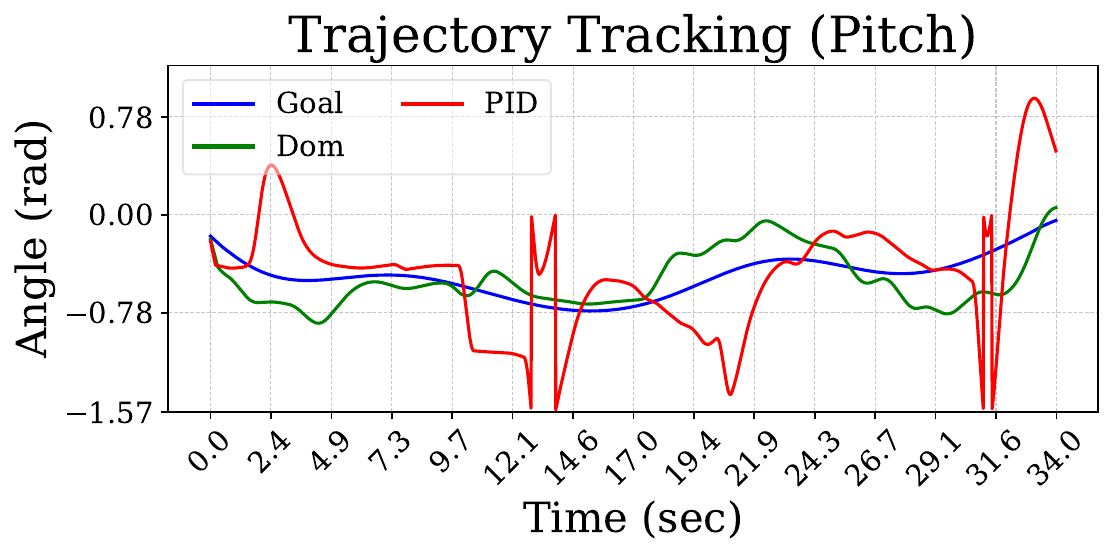}
  \includegraphics[width=0.32\textwidth]{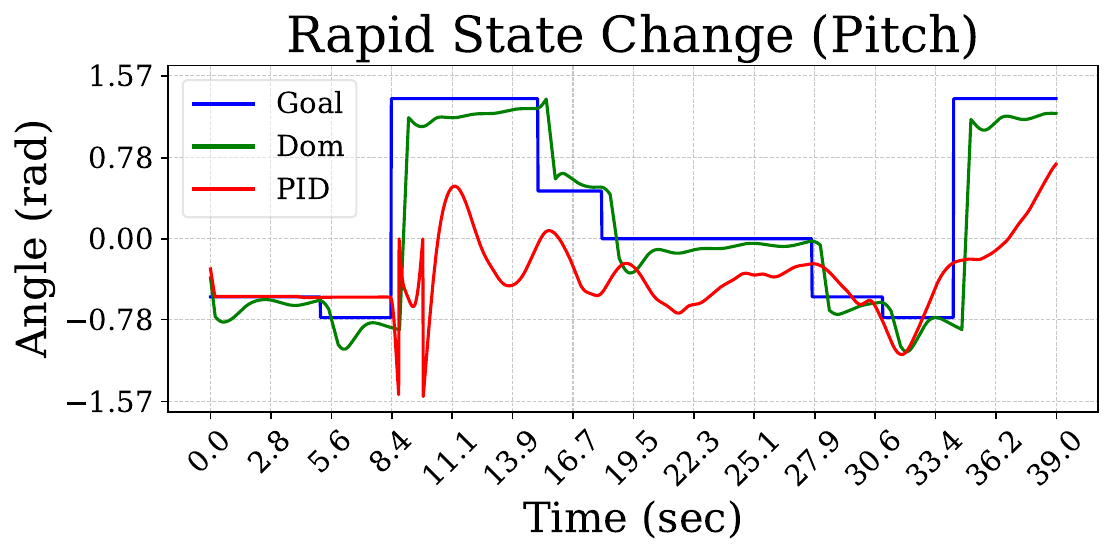}
  \includegraphics[width=0.32\textwidth]{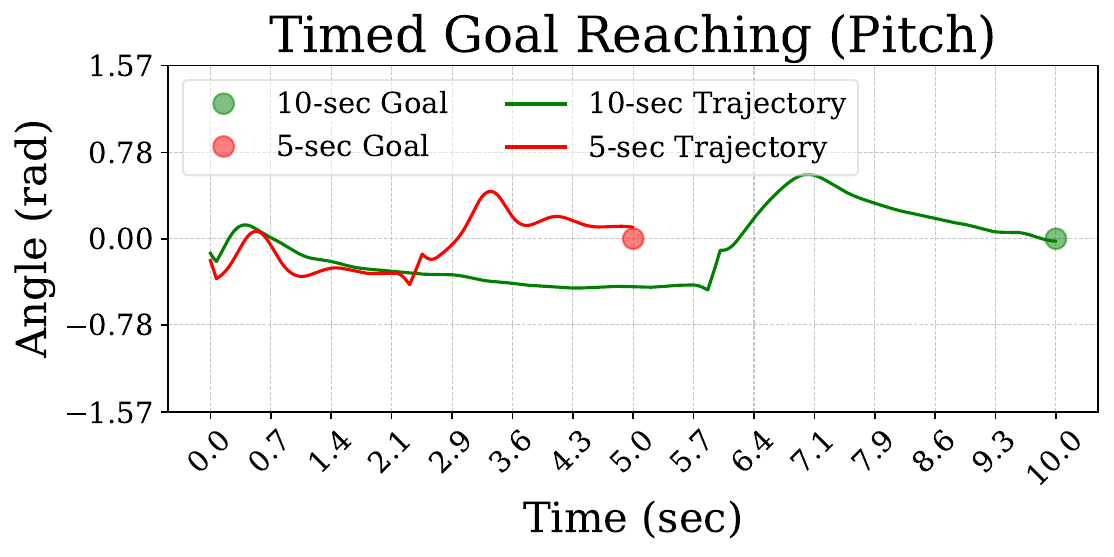}
  \includegraphics[width=0.32\textwidth]{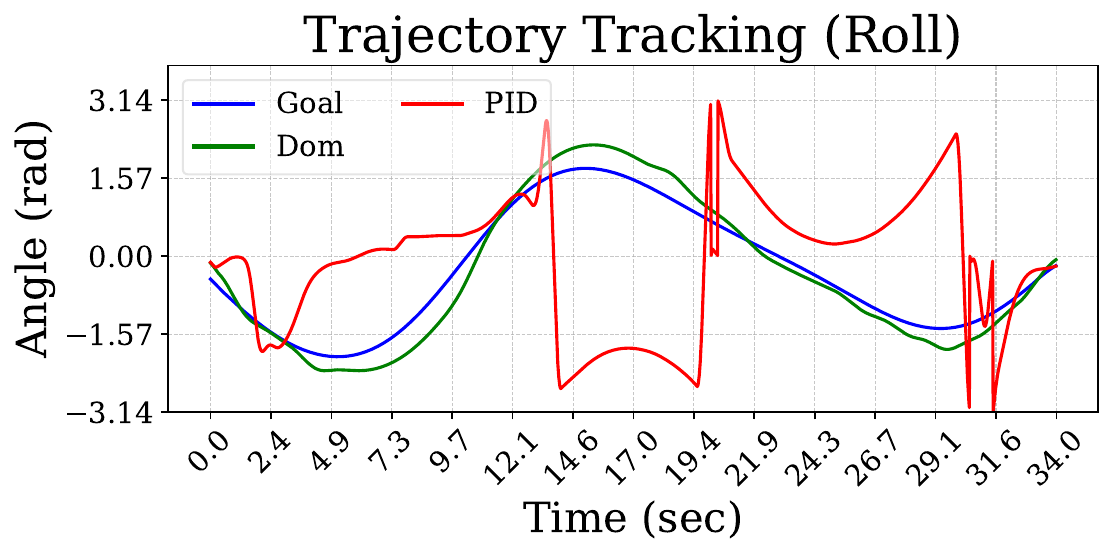}
  \includegraphics[width=0.32\textwidth]{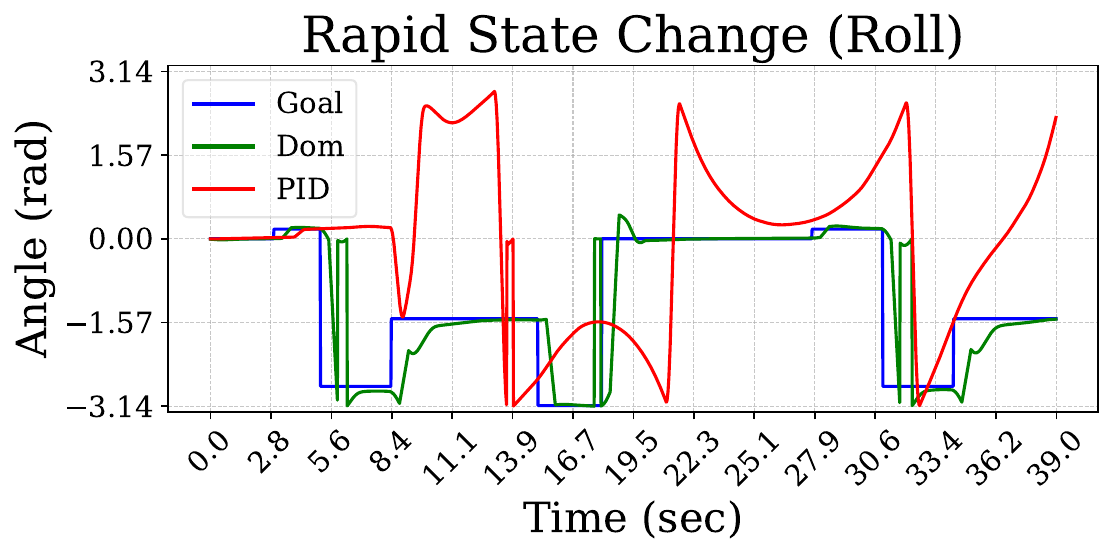}
  \includegraphics[width=0.32\textwidth]{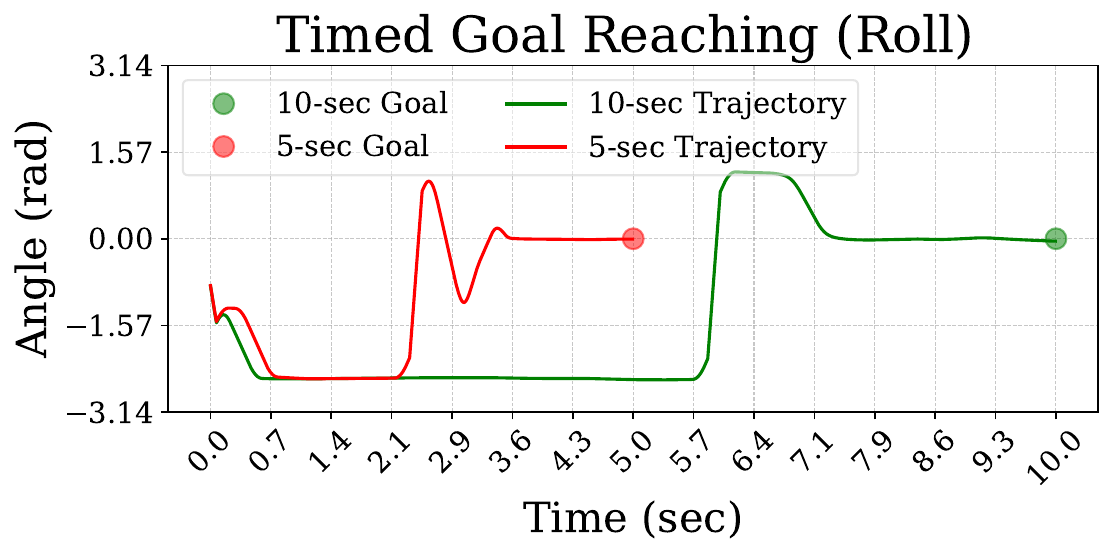}
  \includegraphics[width=0.32\textwidth]{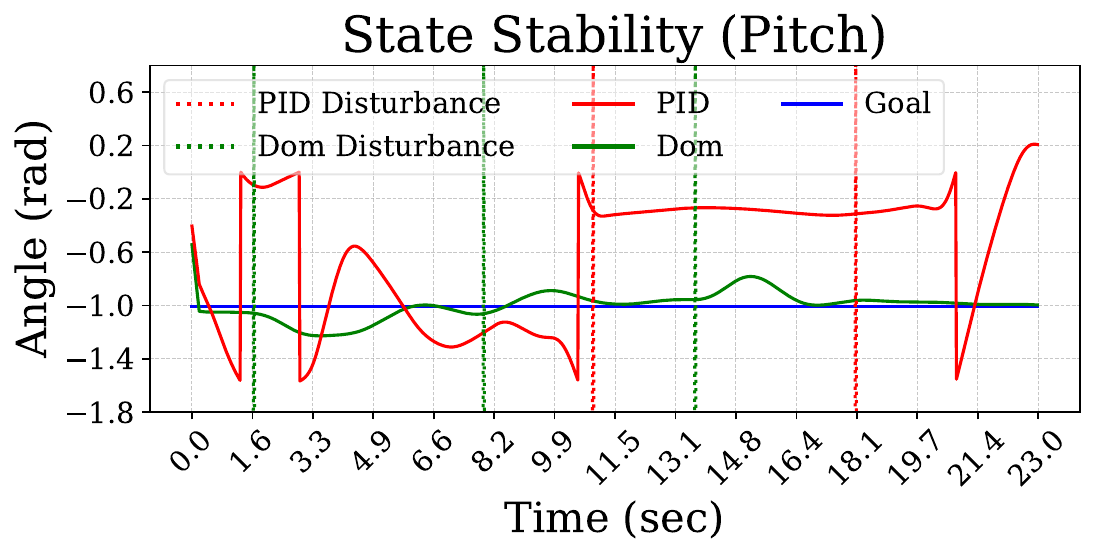}
  \includegraphics[width=0.32\textwidth]{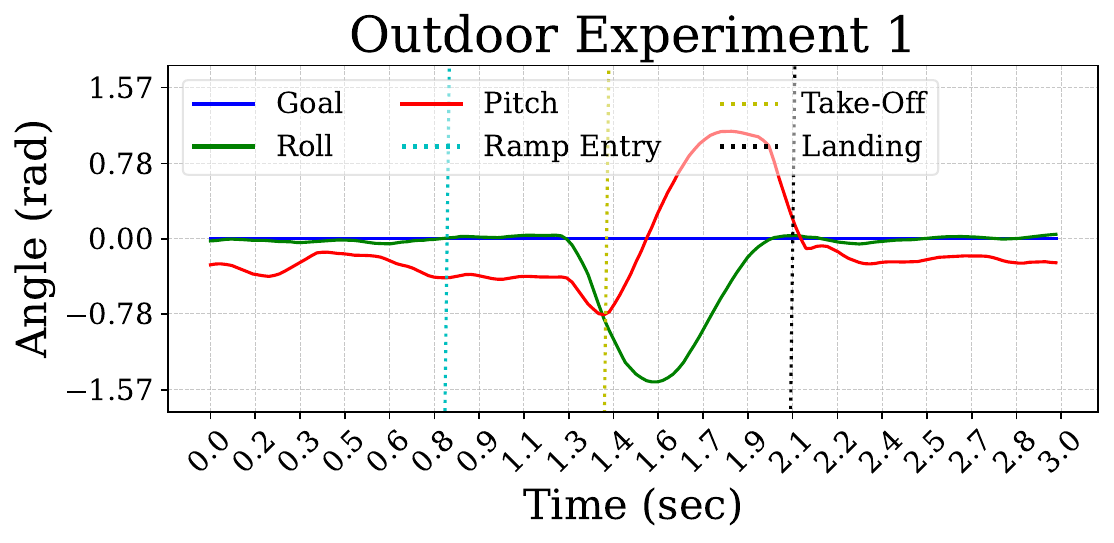}
  \includegraphics[width=0.32\textwidth]{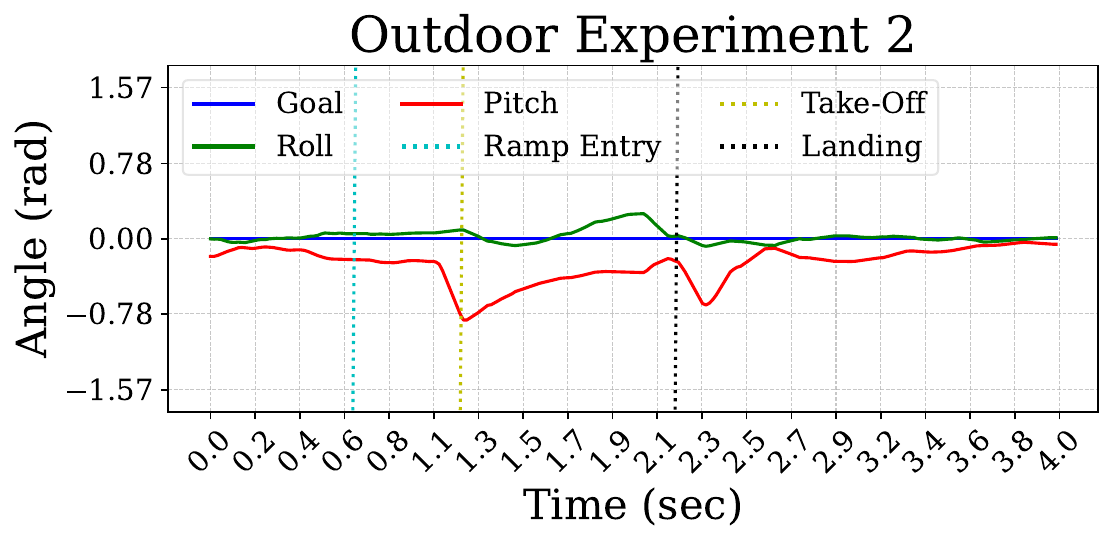}
  \caption{Qualitative Comparison of Roll and Pitch Angle Achieved by \textcolor{green}{\planner} vs \textcolor{red}{PID} w.r.t. \textcolor{blue}{Goal} Angle for Trajectory Tracking, Rapid State Change, Timed Goal Reaching, State Stability, and Outdoor Experiments: \textcolor{green}{\planner}~can closely track continuous \textcolor{blue}{Goal} trajectory (left, top and middle), rapidly achieve discrete random \textcolor{blue}{Goal} states (middle, top and middle), reach timed \textcolor{blue}{Goal} in a timely manner (right, top and middle), keep stable at a constant \textcolor{blue}{Goal} despite external disturbances (left, bottom), and maintain zero pitch and roll during outdoor flights (middle and right, bottom). \textcolor{red}{PID} suffers from poor performance due to a lack of knowledge about coupled vehicle dynamics among state dimensions and actuation limits.}
  \label{fig::error}
\end{figure*}

\subsection{Indoor Experiments}

We design four test scenarios on the gimbal to evaluate different aspects of our approach. Trajectory Tracking (TT) follows a predefined continuous trajectory. Rapid State Change (RSC) switches quickly to a random state and hold it for two seconds. Timed Goal Reaching (TGR) assesses the ability to reach a goal at a defined time. State Stability (SS) assesses the ability to maintain a goal state when facing external disturbances. 
We compare our method against a PID controller as the baseline and report the results in Table \ref{tab::results}. For each experiment, we manually set the initial launch angles, angular velocities, wheel $\textrm{rpm}$, and steering angle $\psi$.

\paragraph{Trajectory Tracking Experiment}
For trajectory tracking, we define a continuous, closed-loop path as a sequence of goal states that involve continuous variations in both roll and pitch. Fig.~\ref{fig::error} (left, top and middle) compares the desired trajectory (blue) with our method (green) and the PID baseline (red). \planner~tracks the desired trajectory closely, while the baseline PID controller deviates significantly and sometimes exceeds state limits, resulting in failure cases where the robot becomes stuck.
We report the mean and standard deviation of the state error, the average time to complete the trajectory (for successful trials), and the overall success rate in Table \ref{tab::results}. 

\paragraph{Rapid State Change Experiment}
In this experiment, the planner randomly selects a new goal state from a predefined set of goal states. After reaching this state, the robot holds it for two seconds before selecting the next random goal. Fig.~\ref{fig::error} (middle, top and middle) show \planner~achieves the new state in a more precise and timely manner than the PID baseline. 
We measure the mean and standard deviation of the time taken to reach each goal and the difference between the achieved state and the goal state along with the overall success rate as reported in Table \ref{tab::results}. 

\paragraph{Timed Goal Reaching Experiment} 
This experiment tests the planner’s ability to arrive at a specified goal state exactly at a predefined time. Fig.~\ref{fig::error} (right, top and middle) qualitatively shows \planner~precisely reaches the goal at the right time (pitch) or reaches the goal earlier and stays there (roll).  We report the mean and standard deviation of the difference between the desired and the actual arrival time and the difference between the achieved final state and the goal state in Table \ref{tab::results}. 

\paragraph{State Stability Experiment} 
In this experiment, the planner first brings the robot to a goal state and holds it for ten seconds.
During the holding phase, we manually disturb the vehicle by applying downward or upward force.
Fig.~\ref{fig::error} (left, bottom) demonstrates the \planner~reacts to three disturbances (vertical dashed lines) and quickly recovers from them, whereas the PID controller struggles to maintain the steady state even without its two disturbances. 
We quantify recovery performance by measuring the mean and standard deviation of the correction time as well as the reaction latency.
As shown in Table \ref{tab::results}, \planner~with \model~consistently stabilizes the system at the goal state faster and with lower reaction latency compared to the PID baseline.

\subsection{Outdoor Experiments}
To validate our method in real-world scenarios, we accelerate the vehicle to about 14 m/s (50+ km/h) and launch it off a ramp set at 45\textdegree~to the ground.
The steering is manually controlled to align the vehicle with the ramp and control is passed to the planner as soon as the vehicle is airborne. 
We estimate time till landing using the equations of projectile motion and pass these inputs to the planner,
which controls the vehicle mid-air for a safe landing at state
$\mathbf{s}_T^g = (0.0, 0.0, 0.0, 0.0, 0.0, 0.0, 1000, 0.0)$, setting wheel $\textrm{rpm}$ to 1000 to reduce impact at landing.
The trials are conducted seven times with different launch poses and velocities. We record the entire trajectory until landing and report the average state upon landing in Table \ref{tab::results}. 
Fig.~\ref{fig::error} (middle and right, bottom) presents two outdoor trials with their corresponding time step for ramp entry, take-off, and landing annotated as vertical dashed lines. In both cases, our \planner~is able to prepare the roll and pitch angles in-air to land the vehicle flat right at the landing time. Considering the safety of the vehicle, we do not conduct outdoor ramp flight experiments with PID. 

\subsection{Discussions}
Based on our experimental observations, while the PID controller can reach the goal state under limited configurations, it often becomes stuck upon exceeding state limits—an issue that can lead to catastrophic failures during landing. In contrast, our \planner~with \model, which leverages forward simulation until the final state with the help of an accurate dynamics model, achieves successful maneuvers by reliably reaching the goal state within the predefined time. 
Future work will focus on integrating the \planner~and \model~into a full off-road navigation system. 

\section{Conclusions}
\label{sec::conclusions}

In this paper, we present the \planner~and \model~to enable in-air vehicle maneuver for high-speed off-road navigation. Based on the precise in-air forward kinodynamics enabled by the hybrid \model~model using physics-based and data-driven approaches, \planner~is able to accurately and timely maneuver the vehicle roll, pitch, yaw, and their velocities to desired states. Extensive experiments showcase that our method allows existing ground vehicle controls, i.e., throttle and steering, to prepare the vehicle in-air for safe landing  during a short airborne period. \textit{So, cars do fly.}

\bibliographystyle{IEEEtran}
\bibliography{mybib}

\end{document}